\documentclass{llncs}
\usepackage{llncsdoc}
\usepackage{amssymb}
\usepackage{fixltx2e}
\usepackage{stfloats}
\usepackage{booktabs}
\usepackage{multicol}
\usepackage{amsmath}
\usepackage[table]{xcolor}
\definecolor{lightgray}{gray}{0.9}
\definecolor{prussian}{rgb}{0.0,0.0,1.0}
\usepackage{multicol}
\usepackage{subfig}
\usepackage{graphicx}
\usepackage{epstopdf}
\usepackage{graphics}
\usepackage{epstopdf}
\usepackage{csquotes}
\usepackage[hypcap]{caption}

\begin{document}
\title{Multi-Agent Shape Formation and Tracking Inspired from a Social Foraging Dynamics}
\author{Debdipta Goswami$^1$, Chiranjib Saha$^2$, Kunal Pal$^3$, Nanda Dulal Jana$^4$ and Swagatam Das$^5$}
\institute{$^{1,2,3}$ Dept. of Electronics and Telecommunication Engineering, \\Jadavpur University, Kolkata\\
$^4$ Department of Information Technology,\\
National Institute Of Technology, Durgapur\\
$^5$ Electronics and Communication Sciences Unit, \\Indian Statistical Institute, Kolkata\\
E-mail: $^1$debdipta\_goswami@yahoo.in, $^2$mail.chiranjib92@gmail.com, $^3$mail.kunalpal@gmail.com,
$^4$nandadulal.jana@it.nitdgp.ac.in and $^5$swagatam.das@isical.ac.in}
\maketitle

\begin{abstract}
Principle of Swarm Intelligence has recently found widespread application in formation control and automated tracking by the automated multi-agent system. This article proposes an elegant and effective method inspired by foraging dynamics to produce geometric-patterns by the search agents. Starting from a random initial orientation, it is investigated how the foraging dynamics can be modified to achieve convergence of the agents on the desired pattern with almost uniform density. Guided through the proposed dynamics, the agents can also track a moving point by continuously circulating around the point. An analytical treatment supported with computer simulation results is provided to better understand the convergence behaviour of the system.
\end{abstract}

\section{Introduction}
In recent years, pattern formation and control of multi-agent autonomous systems has the related research communities. With the tremendous advancement of technology, the interest is also growing in this field where a collection of agents can communicate with each other and also with the environment and can do jobs which are far beyond the capability of a single agent. Primary objective of multi-robot pattern formation is to form a cohesive unit around the region of interest, surveillance and move in a specified disciple. Real-life applications of formation control can be found in multi-robot systems \cite{Buhletal06} \cite{egerstedt01}, micro-satellite clusters \cite{barnesetal09}, unmanned land, sea, or air vehicles \cite{fm04a}, \cite{buzogany}, rendezvous \cite{fm04a}, and mobile robotics \cite{yamaguchi}. 

Common components of a generic shape-formation algorithm can be identified as \cite{Lin}, 1. a homogeneous architecture i.e. components with same hardware and software or heterogeneous with at least one single component having a different configuration, 2. formation control strategy, 3. a behavior based model that describes the nature of the system and environment. Formation control strategy, the most decisive factor of an algorithm can be generally categorized in two primitive classes, centralized and decentralized. In centralized control, the agents follow a single controller that processes all the information while in later, each agent is equipped with its own controller and autonomous decision making capability. Both system has its advantages and disadvantages. Centralized control \cite{ZelinskiKS03}, \cite{1243403}, \cite{coordination}., though is limited by the bandwidth of information exchange or communication between the agents, is more structured for increasing number of robots and complex situations while decentralized system \cite{lawton}, \cite{balch}, \cite{campout}, \cite{leadfollow},  \cite{Lin} having local data processing and less information sharing: no reliability cam be attributed to a single robot or master, falls short when more sophisticated planning is necessary for high level control. Hence, hybrid algorithms have been developed employing both centralized and distributed control compromising the communication cost between the agents to add more robustness to the system \cite{1570486}, \cite{1180220}, \cite{1205192}, \cite{4209428}, \cite{4915742}.

Apart from architectural classifications, the algorithms, by nature can be categorized in three broad classes, namely, leader/neighbor following algorithms, potential field based algorithms, and bio-inspired algorithms. Leader/neighbor-following algorithms \cite{mariottini}, \cite{shao} require the individual robots to follow a trailing neighborhood of the leader or the leader itself, which contains all the global information, eg. the target shape position along with maintaining a specific geometric shape and avoiding collision with each other. This leader-follower approach was modified in \cite{desai02} by adding a control graph that defines the relative position of each robot in the formation. Fredslund and Mataric \cite{fredslund02} used local sensing and minimal communication between agents to preserve a predetermined formation. In a recent approach \cite{pilz}, feedback control and weight adaptation of the interconnect structure of the communication graph topology enhancing the Lyaponov stability has been proposed. The second category of the shape formation algorithms are based on potential field theory \cite{gefua}, \cite{gazi}, \cite{barnesetal09}, \cite{kim06}, \cite{leonard01}, \cite{zelinski}, \cite{fua07}, \cite{tucker}. Typically, a force field is created in the region of interest exerting attractive forces on the agents and the obstacles repelling them, the agents gradually through their directed motion converges to the target shape. Each agent moves along the gradient of the potential field which is the sum of these virtual attractive and repulsive forces. In \cite{gazi}, the authors represented the desired formation pattern in terms of queues and formation vertices. Some artificial potential trenches were employed to represent the desired pattern and trajectory for the group of robots. Although the method greatly improved on node to robot formation structures, queues and vertices were still calculated based on the formation and number of robots. Later a limited communication framework was utilized to improve the work of \cite{gazi} in \cite{fua07}. In \cite{tucker}, a set of artificial points is used as beacons that guide the robots to their goal. This approach is based on the geometric relationship between beacon points to move the robots in formation. In \cite{barnesetal09}, a simple potential field based on scalar sigmoid scaling functions has been proposed where the diverging and converging forces with respect to the origin of a radial coordinate system balances along the target shape and a third curl field exists along the pattern for further stability. The third group of algorithms draw their inspiration from biological systems. Biological systems, ranging from macroscopic swarms of social insects to microscopic cellular systems, can give rise to robust and complex emerging behaviors through much simpler local interactions in the presence of various kinds of uncertainties \cite{kelly}. Several bio-inspired methods for multi-agent shape formation and control have been proposed over the past few years. Motivated by the cell structure, Fukuda \textit{et al.} \cite{fukuda} presented an optimal structure decision method to determine cell type, arrangement, degree of freedom, and link length. They demonstrated that through simulating the cells behavior, multiple agents can form and maintain an optimal structure. Shen \textit{et al.} \cite{Shen} proposed a Digital Hormone Model (DHM) to control the tasking and executing of robot swarms based on local communication, signal propagation, and stochastic reactions. The authors employed Turings reaction diffusion model \cite{turing} to describe the interactions between the hormones. The DHM scheme integrated mechanisms of a dynamic network, stochastic action selection, and hormone reaction-diffusion. Taylor \cite{taylor} proposed a Gene Regulatory Network (GRN) inspired real-time controller for a group of underwater robots to perform a simple clustering task. In that system, the robots can only adapt to local environmental changes without considering its influence on global behaviours. Guo \textit{et al.} \cite{guo} further utilized the GRN inspired dynamics to devise a distributed self-organizing algorithm for swarm robot pattern formation. In \cite{bioinsp}, abstractions of morphogenesis in a network of molecular particles has been employed to manage the spatial self-organization in an ensemble of mobile robots without exploiting the common structural components of shape formation algorithms like global perception, distance and direction sensing intelligence. Apart from all these, few more approaches to formation control by using graph algorithms \cite{desai02}, \cite{fierro02}, models of visual perception \cite{das2002}, model predictive control \cite{dunbarmurry02}, reinforcement learning \cite{kobayashi}, and neural networks \cite{hirota} have also been proposed in literature.

Swarm Intelligence (SI) \cite{bonabeauetal99}, \cite{eberhartetal01} refers to a family of bio-inspired algorithms imitating the collectively intelligent behavior of the groups of natural creatures like bird flocks, fish schools, and insect colonies. One of the natural traits of such communal and cooperative systems is pattern formation. Gravagne and Marks proposed a swarm model \cite{gravange} with very simple rule to update the agents and showed the emergent behaviors of aggressor, protector, and refugee swarms. Andrews \textit{et al.} \cite{andrews} used the social foraging behavior of swarm to design robust multi-agent systems. Particle Swarm Optimization (PSO) \cite{kennedy95}, \cite{delvalleetal08} is one of the leading SI algorithms of current interest. The basic PSO was devised and is still most well-known as a function optimizer in the real-parameter space. In PSO, each trial solution is modelled as a particle and several such particles collectively form a swarm. Particles fly through the multi-dimensional search space following a typical dynamics and searching for the global optima. The PSO dynamics has been improved and rigorously analyzed by several researchers from an optimization point of view. However, very little research work has so far been undertaken to use the dynamics for some purpose like shape formation and tracking, which is completely different from optimization on a functional landscape. At this point we would like to mention a more or less related work in \cite{komann08}, where a swarm of virtual particles, denoted as marching pixels, are crawling according to simple mathematical rules (though these are different from the basic PSO-dynamics) within a 2D pixel image to find the centroids of arbitrary geometrical figures.
 
 In this paper, we propose a simple method for formation of geometric pattern that utilizes a foraging dynamics equipped with a repulsion between nearest neighbours. The intended shape is given as input which we call the target vector to the swarm. The target vector is a parametric vector that defines the shape of the intended pattern and helps the swarm to converge on the desired shape. The stable region of the dynamics is used so that the agents can converge properly. The theoretical results are substantiated using computer simulations.
 
 The organization of the paper is as follows: the section II the proposed system is outlined; in section III the analytical treatment is carried out and the nature of the target vector for each shape is discussed; in section IV the experimental results with simple geometric shapes are shown, the importance of the repulsion between nearest neighbours is explained and the ability of the proposed model for automated tracking is briefly demonstrated. Section V deals with the conclusion and the future possibilities of research.

%
%
%
%
%

\section{Proposed System}
The proposed system for automated shape formation has taken a cue from the foraging dynamics suggested by Das \textit{et al.} in \cite{das2014189}. The basic position update mechanism of the foraging dynamics was given by, 
\begin{equation}
\dot{\vec{x_i}}=\sum_{j=1,j\neq i}^{M}\dfrac{k(\vec{x_j}-\vec{x_i})}{[\sigma ^2 + \Vert \vec{x_j}-\vec{x_i} \Vert ^2]^\beta} + \dfrac{k(\vec{p}-\vec{x_i})}{[\sigma ^2 + \Vert \vec{p}-\vec{x_i} \Vert ^2]^\beta}\;\;\;,
\end{equation}
where $\vec{p}\in\Re^n$ is the optimum of the artificial potential field considered to simulate the real world application. Here the swarm is being pushed or attracted towards $\vec{p}$ to achieve the desired goal. For shape formation purpose we want to modify this, dynamics so that the agents converge on a pre-defined shape with uniform spacing. $k$, $\sigma$ and $\beta$ are positive constant parameters as defined in \cite{das2014189}.
 One basic requirement for this is to attract each particle towards the boundary of the shape that we are intended to form. It is also desired that the particles will be equally spaced along the shape and remain uniformly separated from each other as much as possible. Hence, each particle should repel its nearest neighbour. Thus the directing equation of the swarm that helps to form different shapes are given by,
\begin{equation}
\begin{split}
\dot{\vec{x_i}}=\sum_{j=1,j\neq i}^{M}\dfrac{k_1(\vec{x_j}-\vec{x_i})}{[\sigma ^2 + \Vert \vec{x_j}-\vec{x_i} \Vert ^2]^\beta} + \dfrac{k_2(\vec{x_T}-\vec{x_i})}{[\sigma ^2 + \Vert \vec{x_T}-\vec{x_i} \Vert ^2]^\beta}\\
+r\dfrac{\vec{x_i}(t)-\vec{x_n}(t)}{\Vert \vec{x_i}(t)-\vec{x_n}(t)\Vert}\;.
\end{split}
\end{equation}
where $\vec{x_i}(t)$ is the position vector of an individual in the swarm. $\vec{x_T}(t)$ denotes the target vector which controls the motion of the individual and depends on the geometric structure to be formed and the position of the individual at time $t$. $\vec{x_n}(t)$ is the position vector of the nearest individual of $\vec{x_i}(t)$ ; $||\bullet||$ denotes the Euclidean distance function; $k_1$, $k_2$ and $r$ are constants.

This first term is inspired from the mutual interaction of the agents in foraging dynamics, the second term is responsible to form a geometric structure based on an attraction. The third term is a mutual repulsion term introduced to ensure that the agents of the swarm do not agglomerate or collide with each other and thus maintaining a uniform density over the entire shape. If the repulsion term is absent, the swarm will still form the desired structure, however, the symmetry in the structure will be very poor. This has been illustrated in Section 4 through computer simulations.

\section{Analytical Treatment}
The foraging dynamics proposed by \cite{das2014189} has distinct damped, limit-cyclic and chaotic behaviour in its stable region. For shape formation, it's important that the agents will converge in the desired formation quickly. So the damped oscillatory behaviour is more appropriate to use. The aparameters of the dynamics have been so chosen that it remains within the limit of stable damped oscillatory behaviour.

To form different shapes, different types of target vectors are necessary accordingly. In this section we discuss about the target-vectors needed to produce the patterns common in literature.

\subsection{Straight Line}
For forming a straight line with equation: $x_1=mx_2+c$, the target vector $\vec{x_T}(t)$ is
given by:
\begin{equation}
\vec{x_T}(t)=\biggl[\dfrac{(x_1+mx_2)-mc}{1+m^2},\dfrac{m(x_1+mx_2)+mc}{1+m^2}\biggr]^T.
\end{equation}
For simulation purpose we use the following values to ensure the stable behaviour of the dynamics: $k_1=0.1,k_2=2.0,\sigma=3.5, \beta=1.2, m=1, c=1$. The initial values for the positions are taken randomly between -5 and 5. The solution for a single agent is shown in Figure~\ref{fig::osc_st_line}.
\subsection{Ellipse}
To form an ellipse with the equation $\dfrac{x_1^2}{a^2}+\dfrac{x_2^2}{b^2}$, the target vector needs to be \begin{equation}
\vec{x_T}(t)=[a\cos\theta,b\sin\theta]^T,
\end{equation} where $\theta=tan^{-1}\left(\dfrac{ax_2}{bx_1}\right)$. The swarm parameter values are kept same as that of the straight line case and the ellipse parameter values are set at $a=4$ and $b=2$.
\subsection{Circle}
A circle is a special kind of ellipse where $a=b=r_0$ and $x_1^2+x_2^2=r_0^2$, , being the equation of the circle, the target vector can be simply written as:
\begin{equation}
\vec{x_T}(t)=[r_0\cos\theta,r_0\sin\theta]^T,
\end{equation} where $\theta=tan^{-1}\left(\dfrac{x_2}{x_1}\right)$. In this case $r_0$ is taken as $4$. The plot for a single agent's position w.r.t. time is shown in Figure~\ref{fig::osc_circle}.
\begin{figure*}[ht]
        \centering
        \subfloat[]{
                \includegraphics[width=0.24\textwidth]{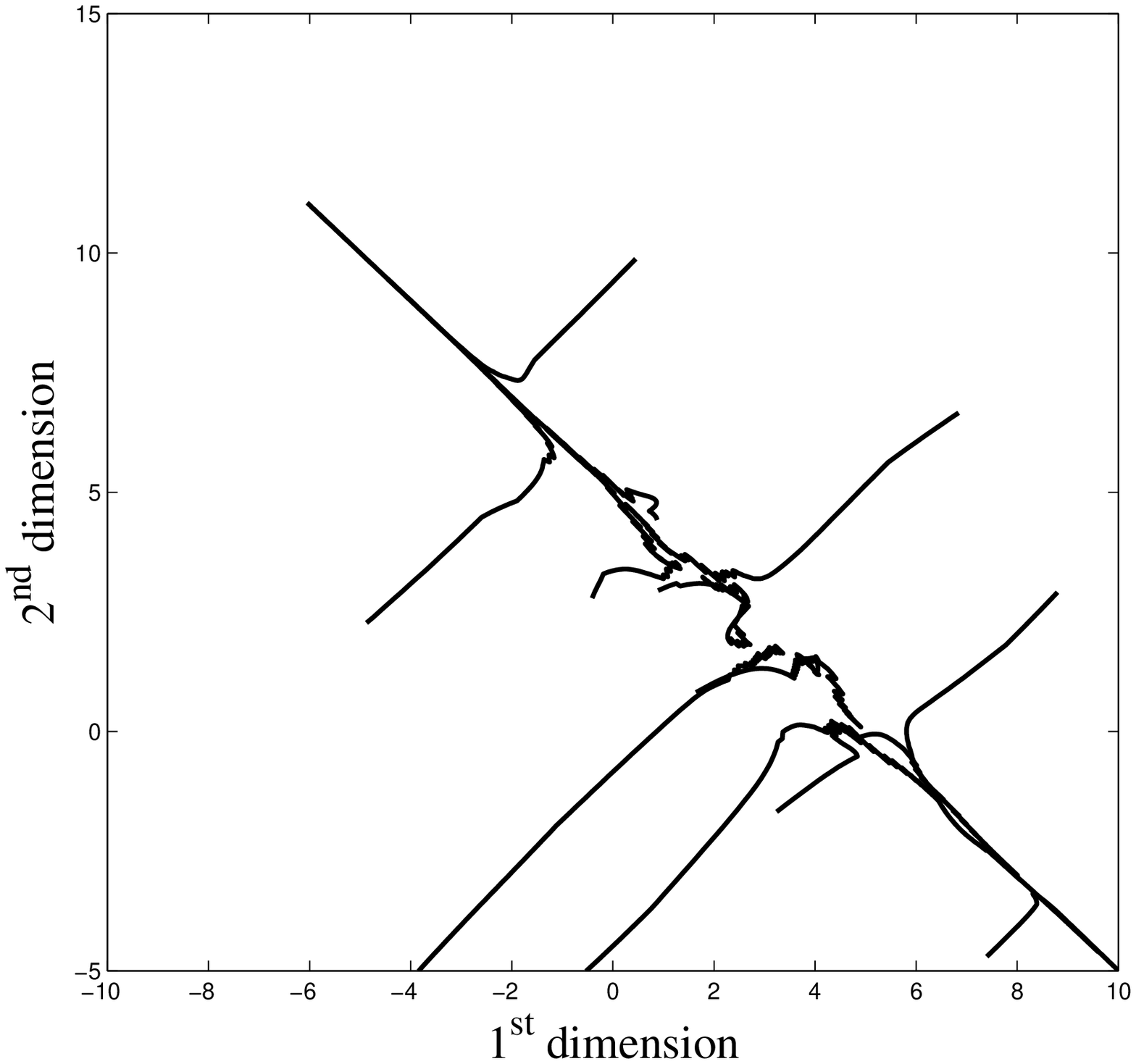}
                \label{fig:st.line.1}
        }
        \subfloat[]{
                \includegraphics[width=0.24\textwidth]{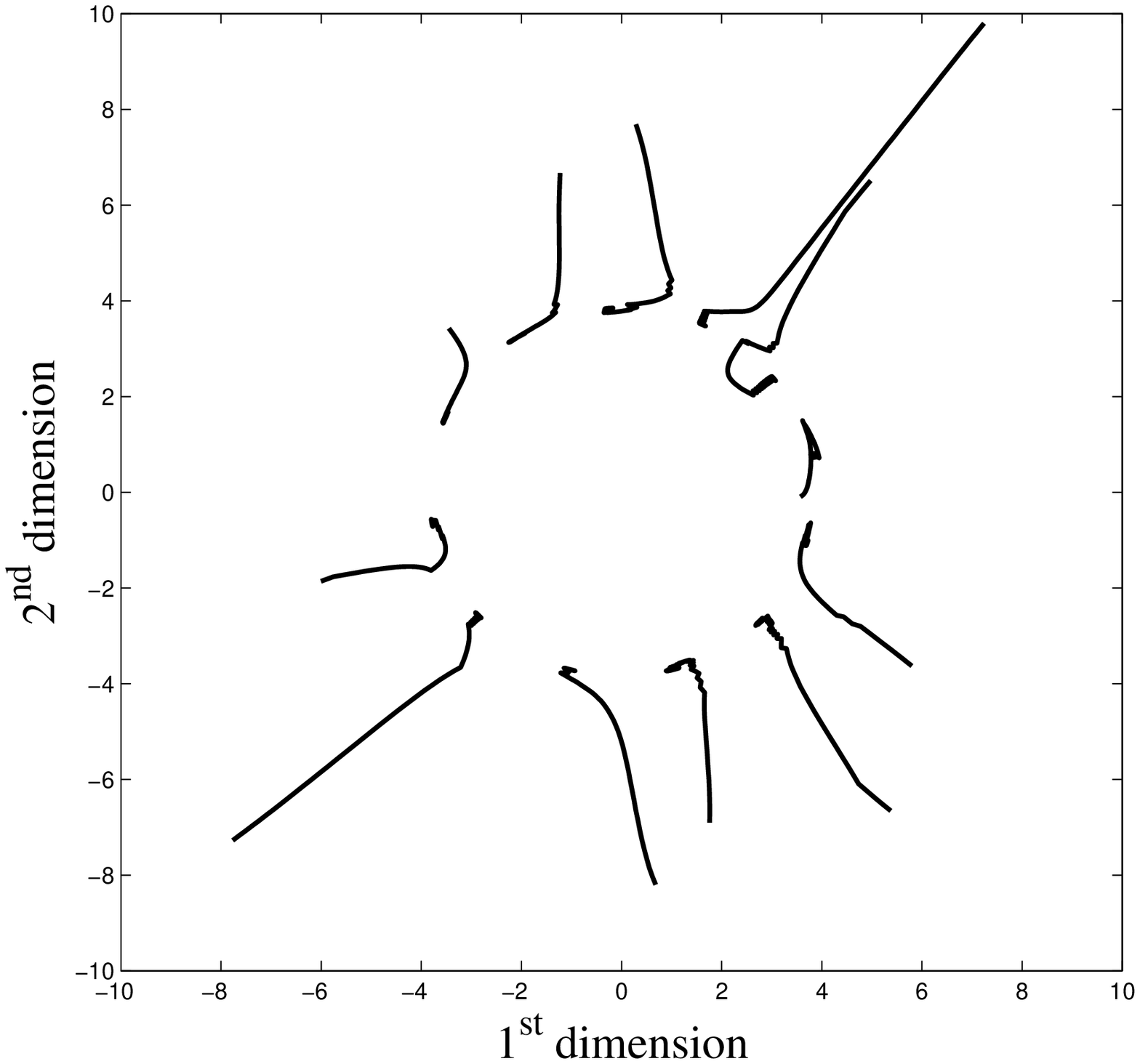}
                \label{fig:cicrle.1}
        }
        \subfloat[]{
                \includegraphics[width=0.24\textwidth]{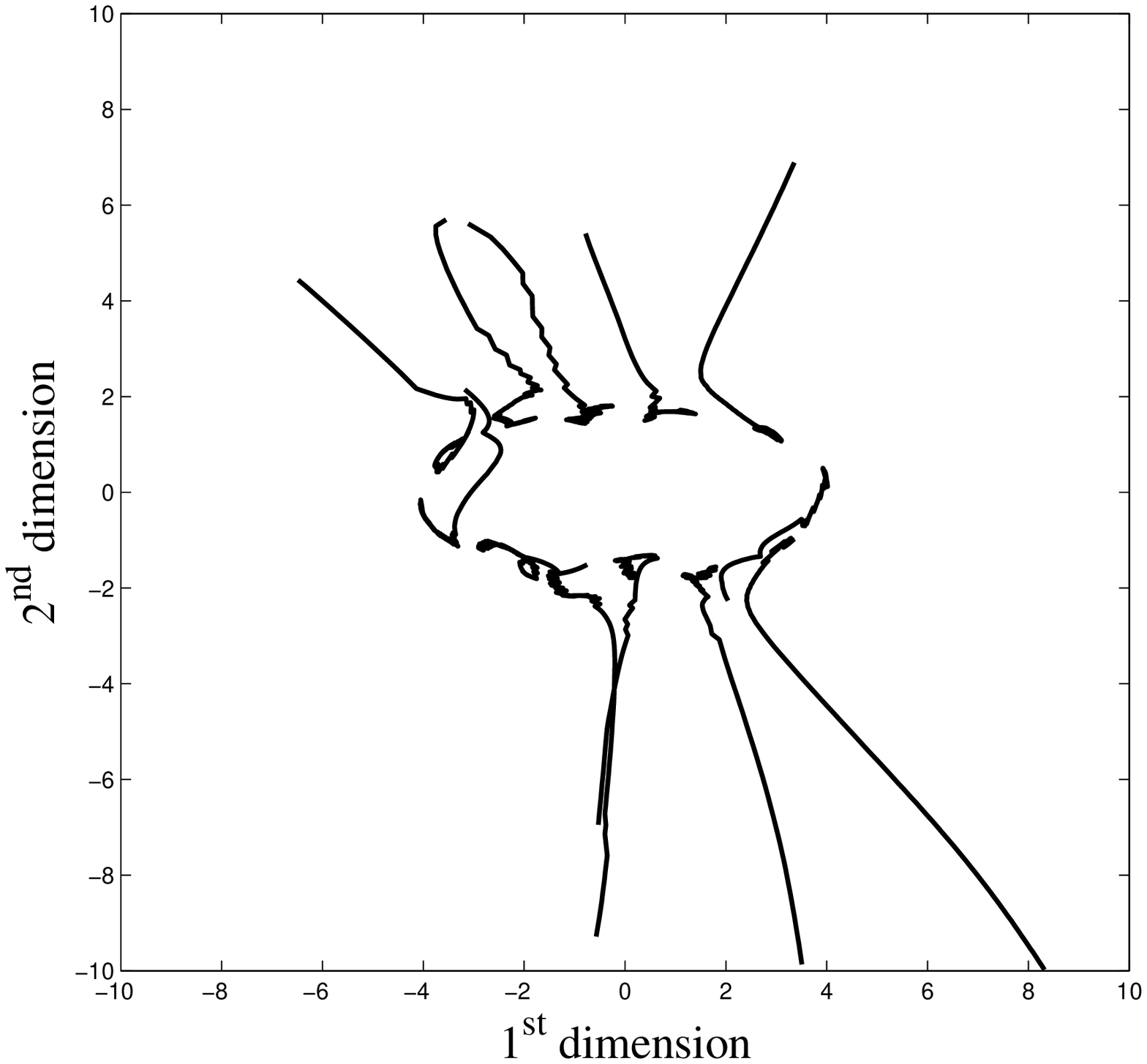}
                \label{fig:ellipse.1}
        }
        \subfloat[]{
                \includegraphics[width=0.24\textwidth]{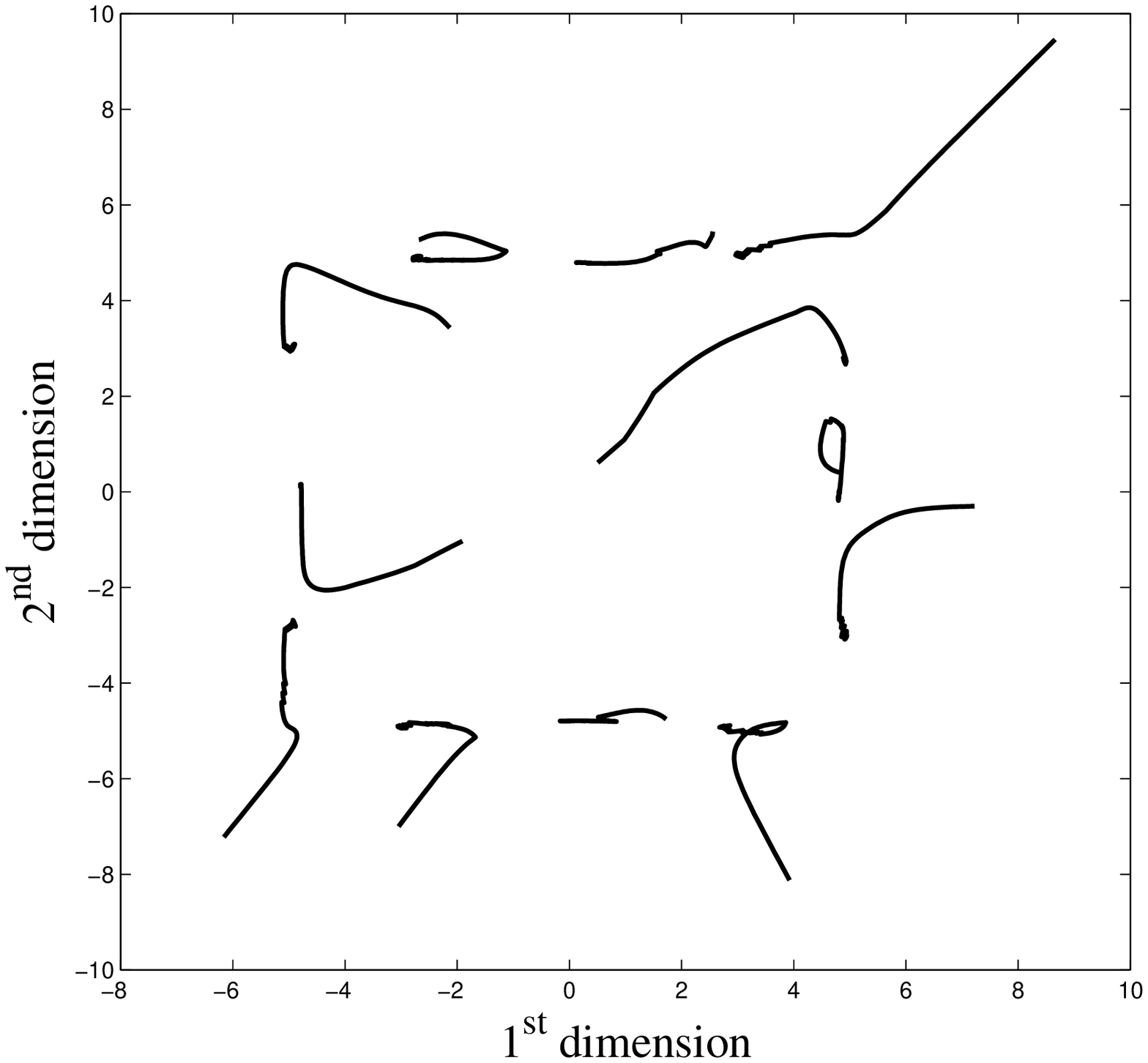}
                \label{fig:rectangle.1}
        }
        \vfill
        \centering
        \subfloat[]{
                \includegraphics[width=0.24\textwidth]{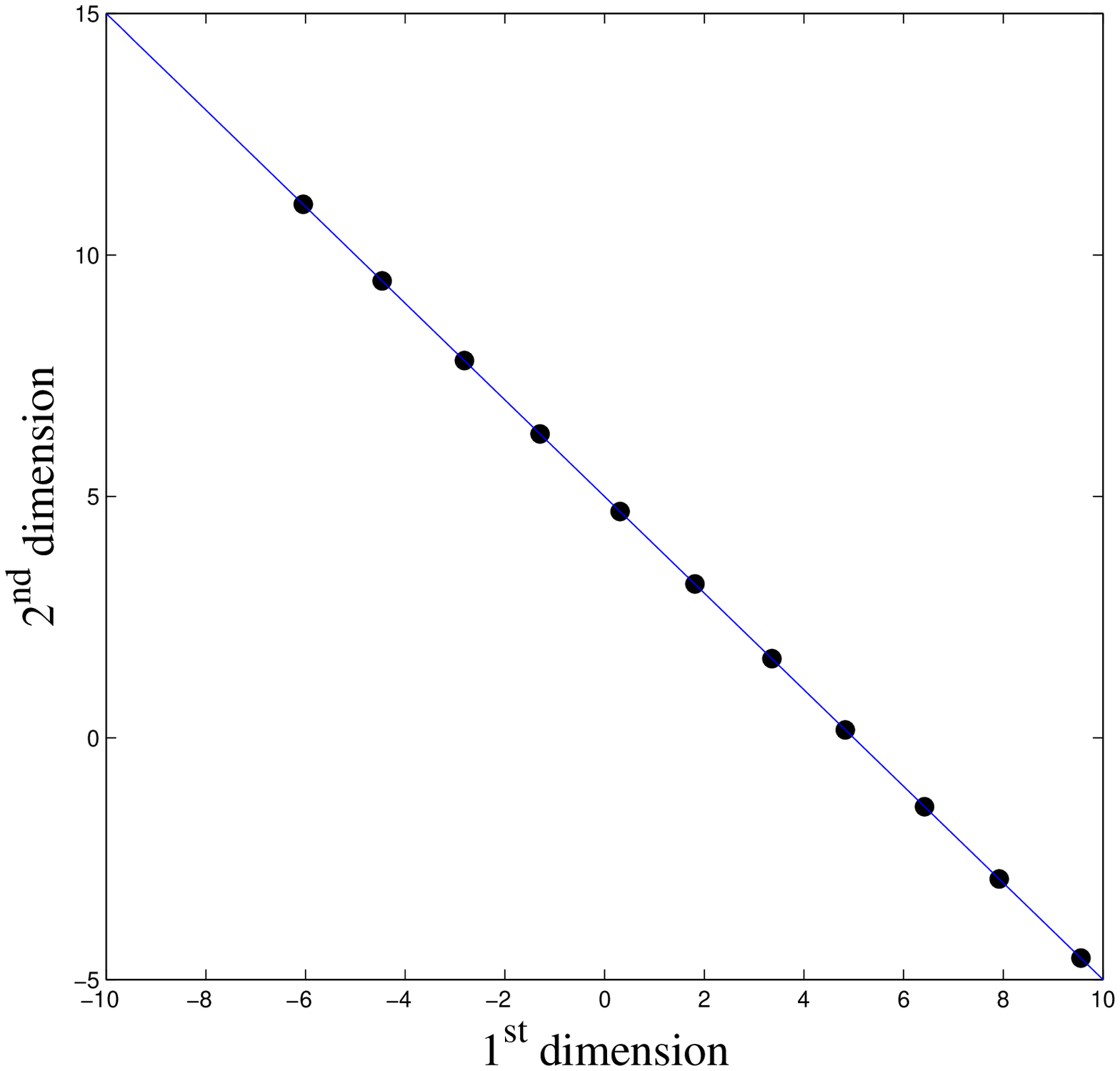}
                \label{fig:st_line.2}
        }
        \subfloat[]{
                \includegraphics[width=0.24\textwidth]{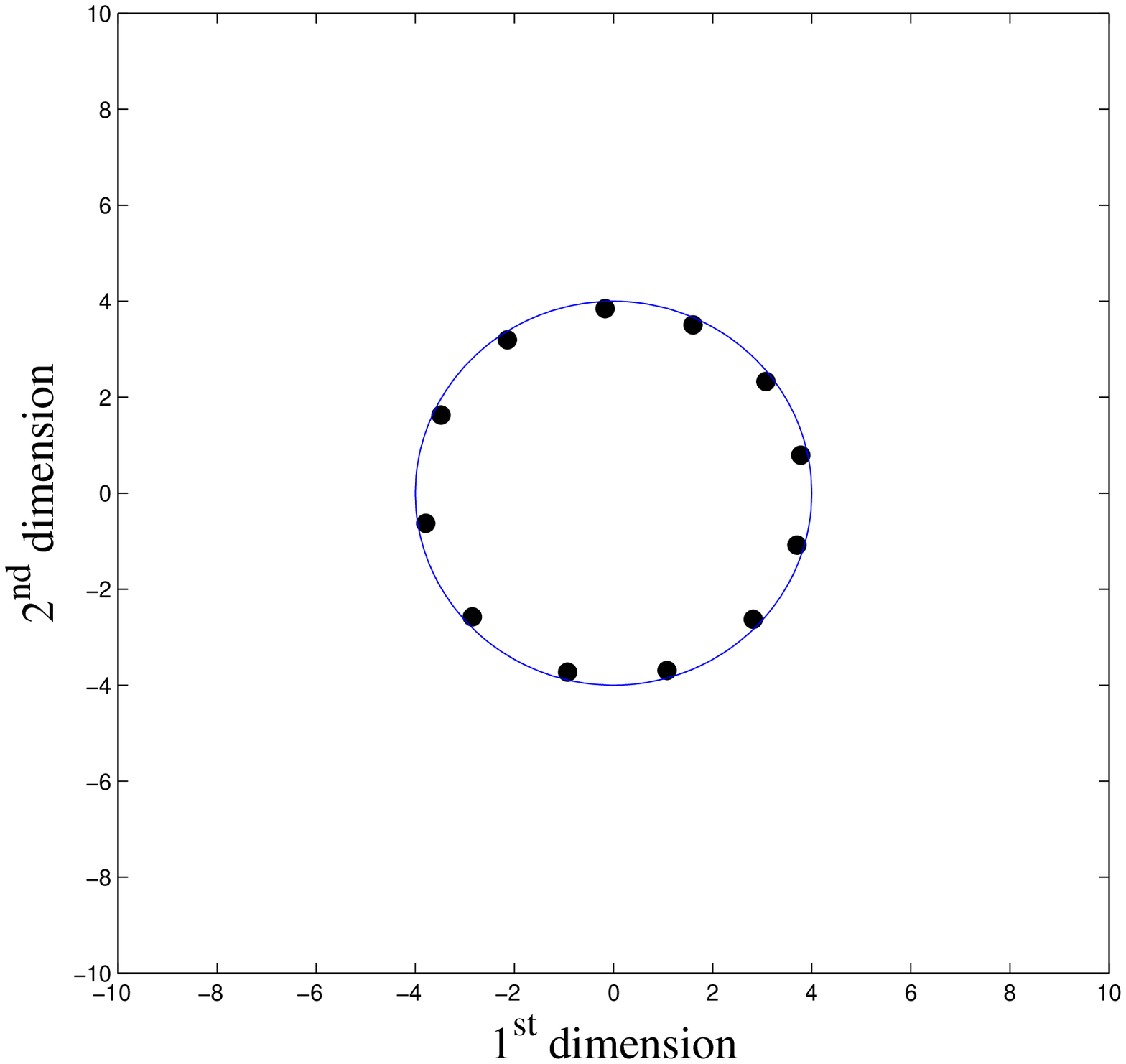}
                \label{fig:circle.2}
        }
        \subfloat[]{
                \includegraphics[width=0.24\textwidth]{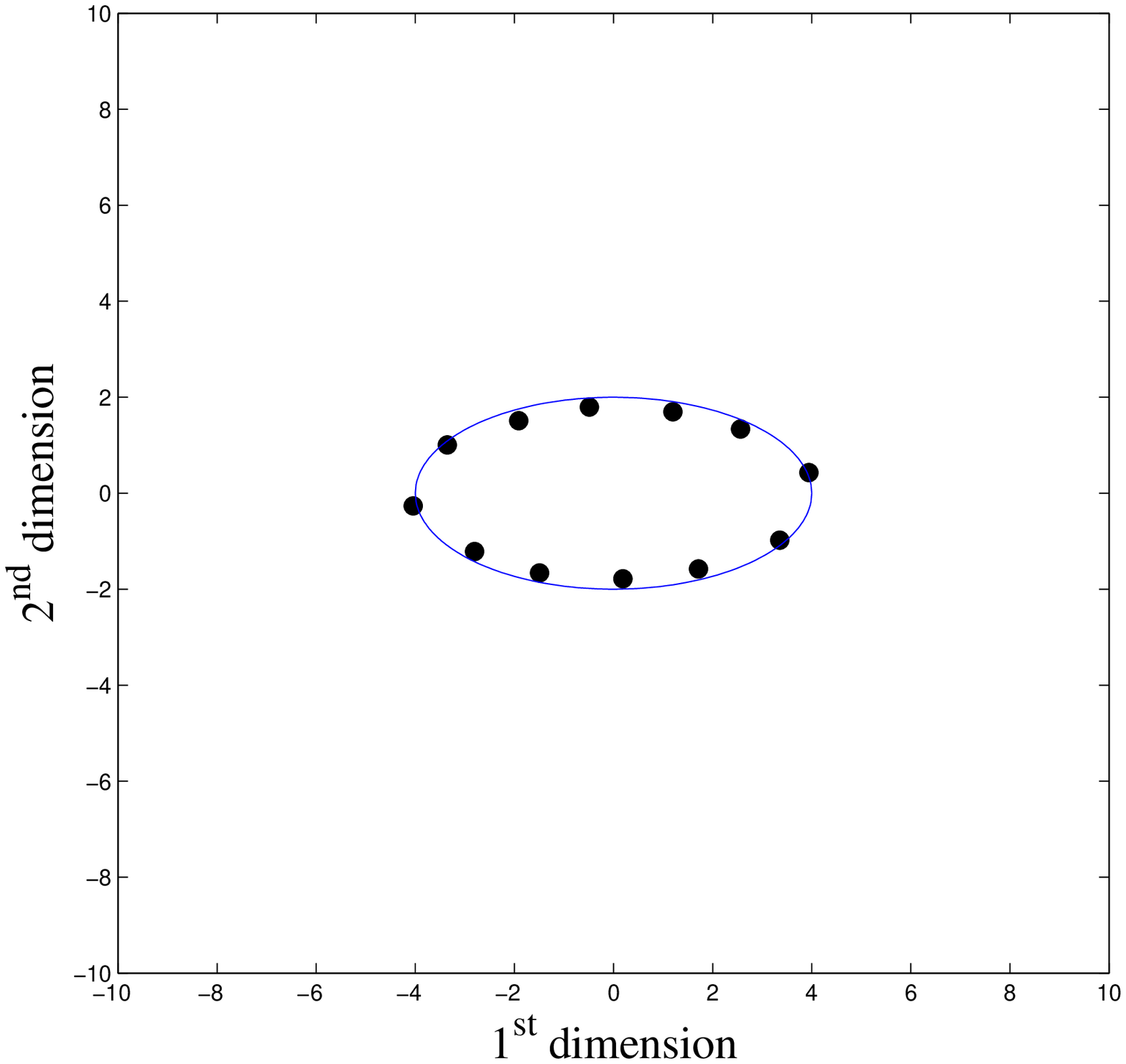}
                \label{fig:ellipse.2}
        }
        \subfloat[]{
                \includegraphics[width=0.24\textwidth]{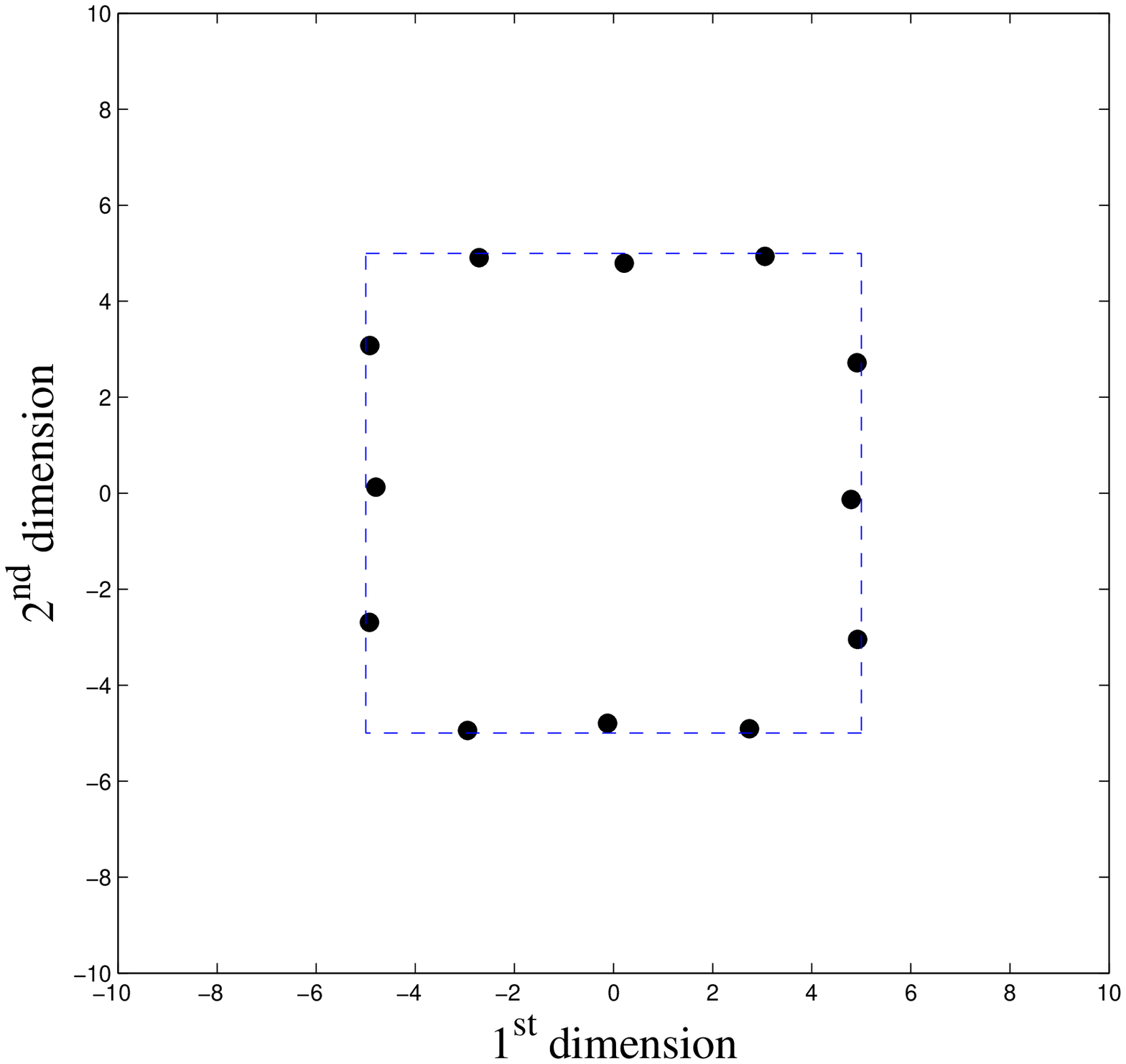}
                \label{fig:rectangle.2}
        }
        \caption{\ref{fig:st.line.1}-\ref{fig:st_line.2} Formation of a straight line; \ref{fig:cicrle.1}-\ref{fig:circle.2} formation of a circle; \ref{fig:ellipse.1}-\ref{fig:ellipse.2} formation of an ellipse; \ref{fig:rectangle.1}-\ref{fig:rectangle.2} formation of a rectangle.}\label{fig:Large_figure}
       
\end{figure*}

\subsection{Square}
\begin{figure*}[ht]
       
        \centering
        {
        \subfloat[]{
         \includegraphics[width=0.31\textwidth]{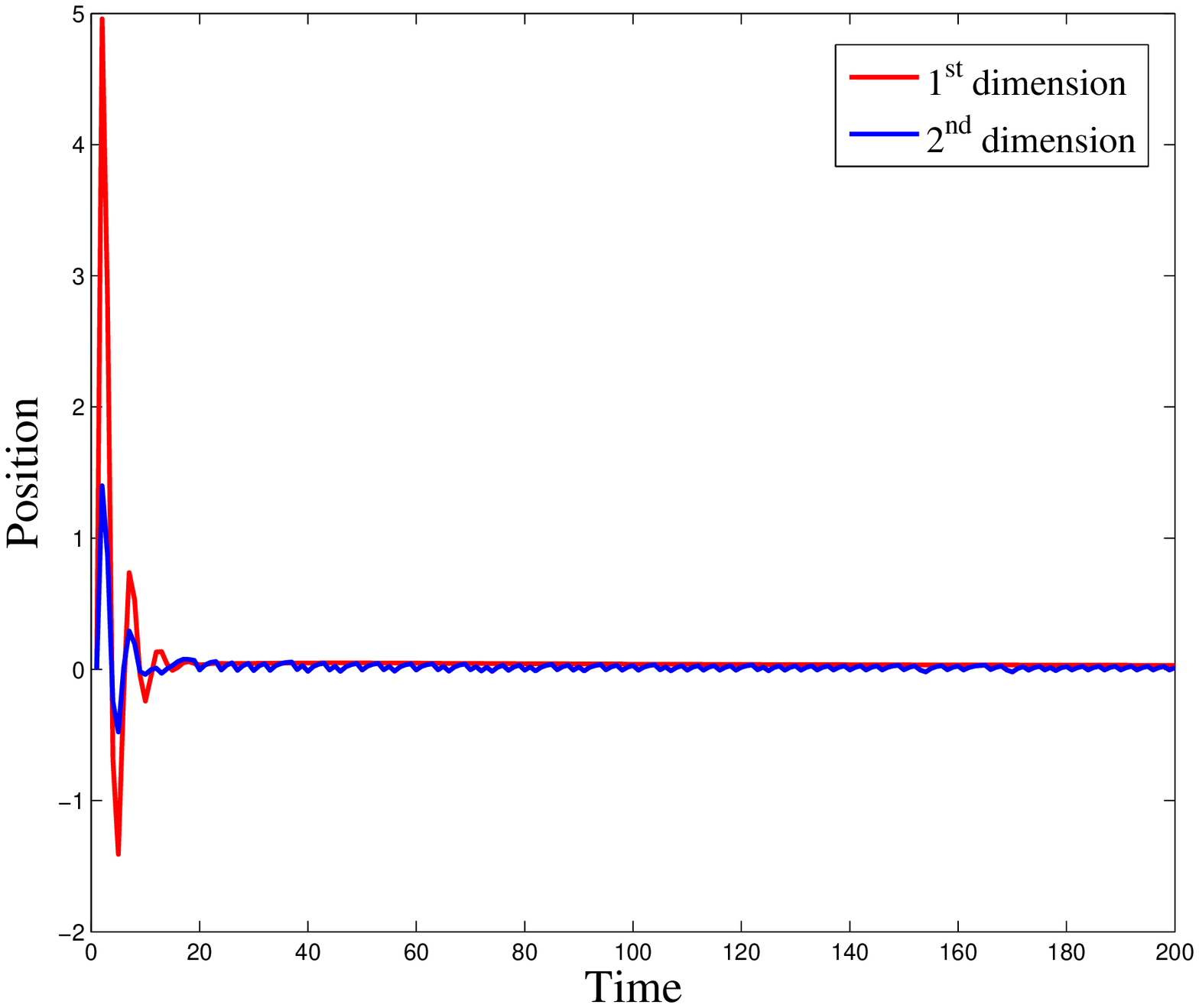}  
       \label{fig::osc_st_line}
        }          
        \subfloat[]{        		\includegraphics[width=0.31\textwidth]{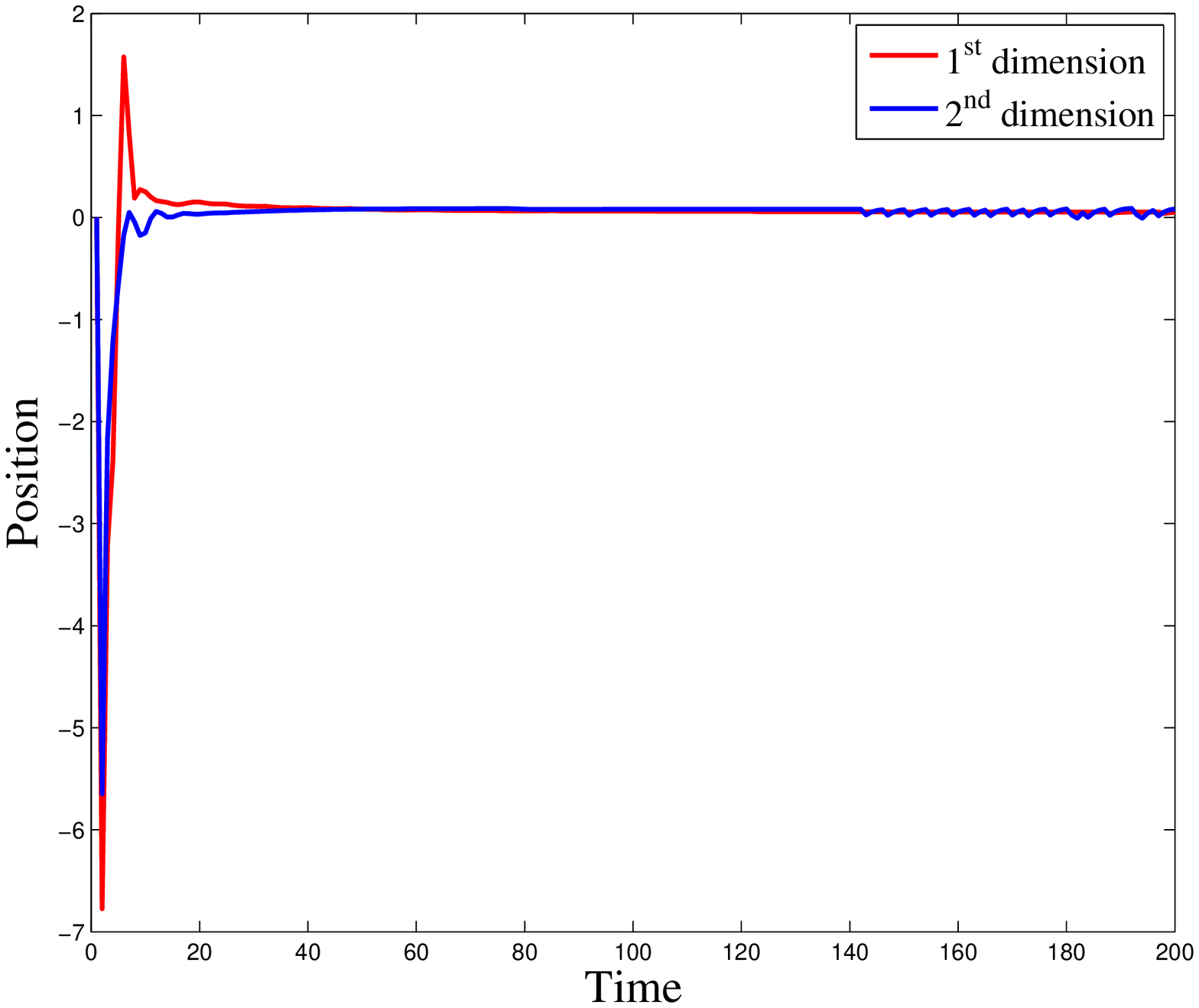}
                \label{fig::osc_circle}
        }
        \subfloat[]{
                \includegraphics[width=0.31\textwidth]{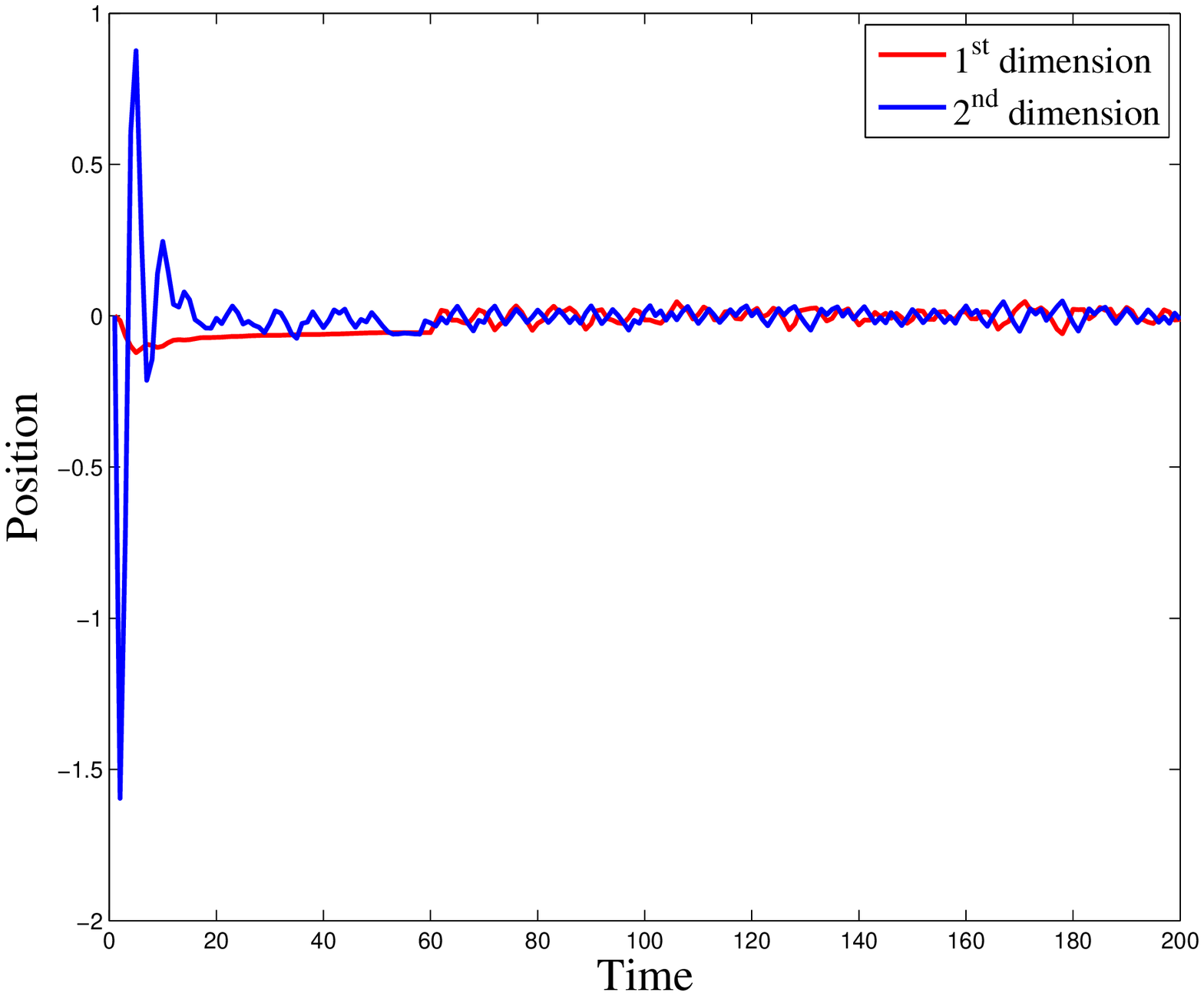}
                \label{fig::osc_rectangle}
        }
       }
        \caption{Variation of position of an agent with respect time for formation of straight line (\ref{fig::osc_st_line}), circle (\ref{fig::osc_circle}) and rectangle (\ref{fig::osc_rectangle}).
a circle}\label{fig::osc_Large_figure}
       
\end{figure*}

We consider a simple square with length of side 2a, the sides are parallel to the coordinate axes and the square is centered at origin. Mathematically it can be represented by two pair of parallel straight lines $|x_1|=a$ and $|x_2|=a$ within the ranges $|x_1|<a$ and $|x_2|<a$ respectively. In this case, the target vector can be written as follows:
\begin{eqnarray}
\vec{x_T}(t)=[a,ma]^T, & for -\dfrac{\pi}{4}<\phi<\dfrac{\pi}{4}\\
\vec{x_T}(t)=[\dfrac{a}{m},a]^T, & for \dfrac{\pi}{4}<\phi<\dfrac{3\pi}{4}\\
\vec{x_T}(t)=[-\dfrac{a}{m},-a]^T, & for -\dfrac{3\pi}{4}<\phi<-\dfrac{\pi}{4}\\
\vec{x_T}(t)=[-a,-ma]^T & otherwise;
\end{eqnarray}
where $m=\dfrac{x_2}{x_1}$ and $\phi$ is the angle obtained when $[x_1,x_2]$ is transformed into polar coordinates. The simulated position coordinates of a single agent is shown in Figure~\ref{fig::osc_rectangle} while forming a square.

\section{Experimental Results}
\subsection{Formation of Simple Patterns}
The proposed method is used to form simple structures like straight-lines, ellipses, circles and squares. For that purpose we have fixed our parameter values in the stable region of the dynamics as $k_1=0.1,k_2=2.0,\sigma=3.5, \beta=1.2$ and $r=0.1$. Formation specific parameters are set as: m = −1, c = 5 for
straight line; $a=4, b=2$ for ellipse and $r_0=4$ for circle. For the case of formation of square, the used parameter a = 5. The final formations are presented in 	Figures~\ref{fig:st.line.1}-~\ref{fig:rectangle.2}. The final
positions of the agents are shown as the tiny filled circles and the targeted geometrical formations are drawn in dashed lines.
\subsection{Usefulness of the repulsion term}
\begin{figure}[ht]
        \centering
        \subfloat[Ellipse]{
                \includegraphics[width=0.45\textwidth]{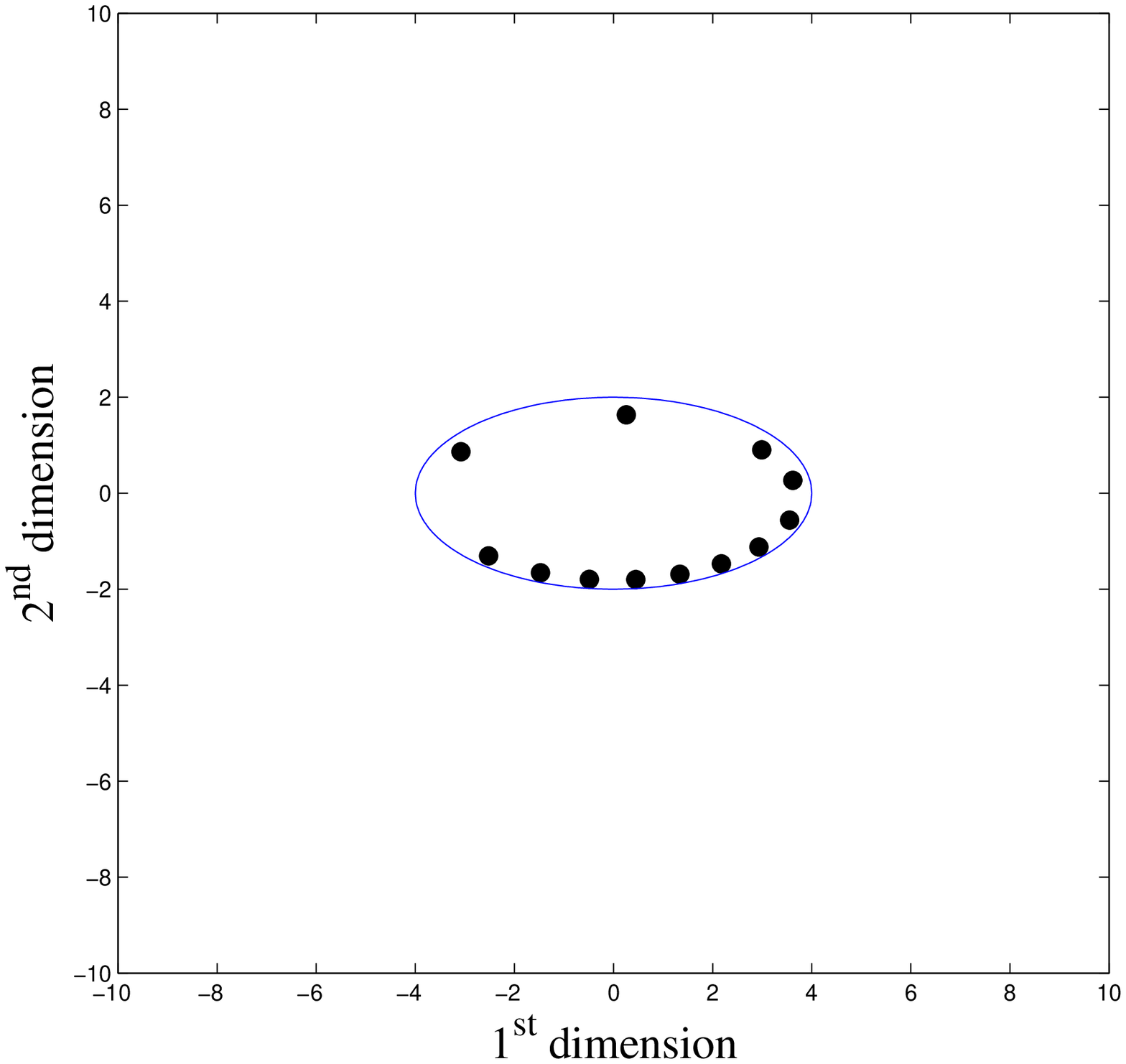}
                \label{fig:ellipse_wo_repulsion}
        }%
        \subfloat[Rectangle]{
                \includegraphics[width=0.45\textwidth]{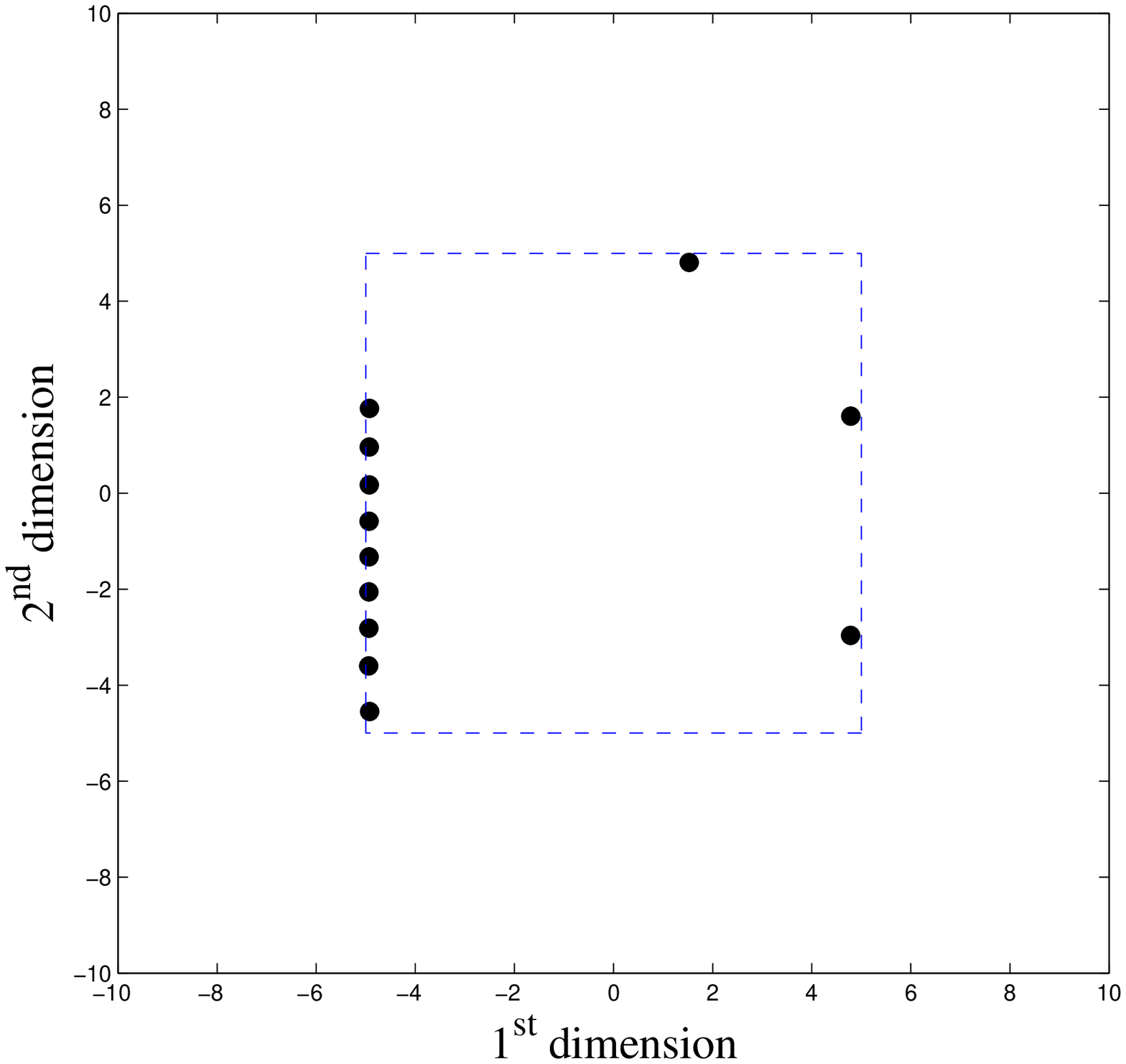}
                \label{fig:rectangle_wo_repulsion}
        }
        \caption{Shape formation without repulsion}\label{fig:shape_wo_repulsion}
\end{figure}

The usefulness of the repulsion term in equation (2) is also investigated. For this purpose, we again simulate the algorithm with the parameter settings $r=0$ and in the
other case $r=0.1$; keeping the other parameters fixed at  $k_1=0.1,k_2=2.0,\sigma=3.5$ and $\beta=1.2$, for a
12 agent swarm forming an ellipse and a square. The final formed structures are shown in
Figure~\ref{fig:shape_wo_repulsion}. In the first case, the repulsion term is absent, hence the agents do not communicate with each other and the final formation lacks from the problem of symmetry which can be overcome by introducing the repulsion term. This can be observed from
Figure~\ref{fig:shape_wo_repulsion} very well.
\subsection{Tracking Ability}

The proposed system can be used to track a moving point in real-time by surrounding it. This important property of continuously surrounding a moving point can be put to use in real life in case of the automatic protection of a convoy by swarm robots. The formation of circle can be used in this case for surrounding the moving point and the moving point can be treated as the center of the circle to be formed. To accomplish this the target vector has to be modified slightly as, 
\begin{figure}[t]
        \centering
        \subfloat[]{
                \includegraphics[width=0.45\textwidth]{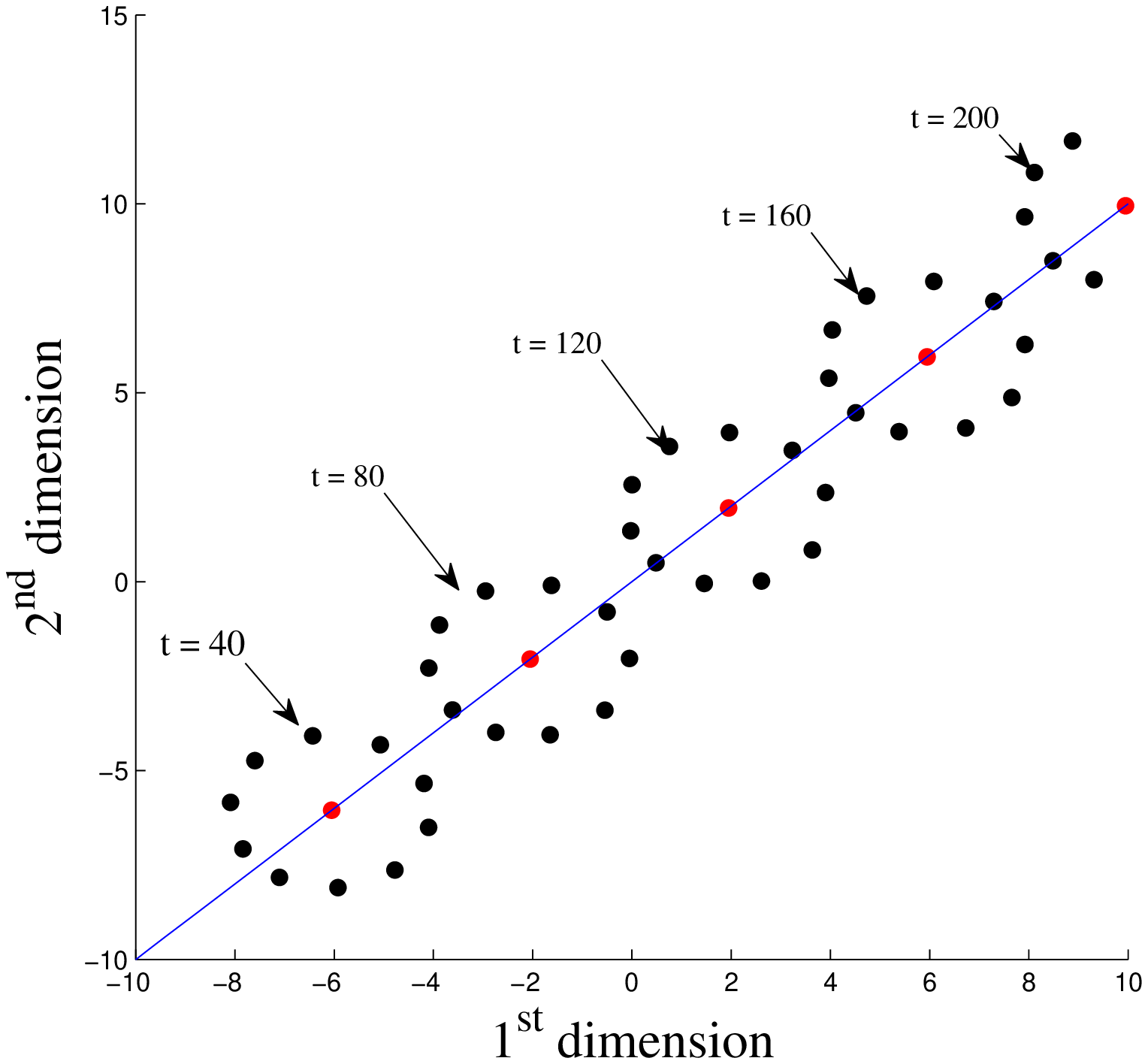}
                \label{track_st_line}
        }%
        \subfloat[]{
                \includegraphics[width=0.45\textwidth]{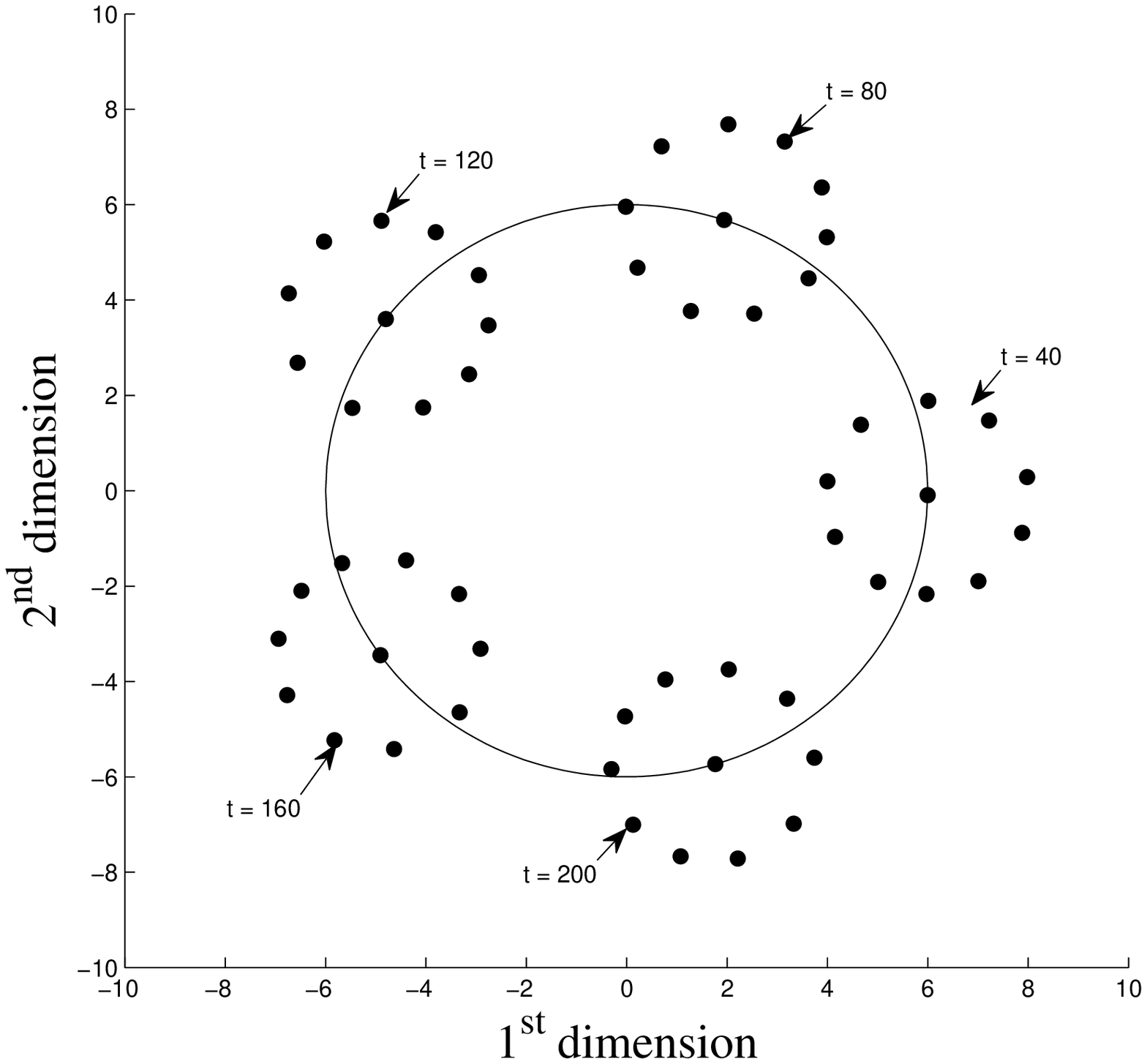}
                \label{track_circular}
        }
        \caption{Circle formed around the center which is moving in a straight line (\ref{track_st_line}) and circle (\ref{track_circular}). }
\end{figure}
\begin{figure}[ht]
        \centering
        \subfloat[]{
                \includegraphics[width=0.45\textwidth]{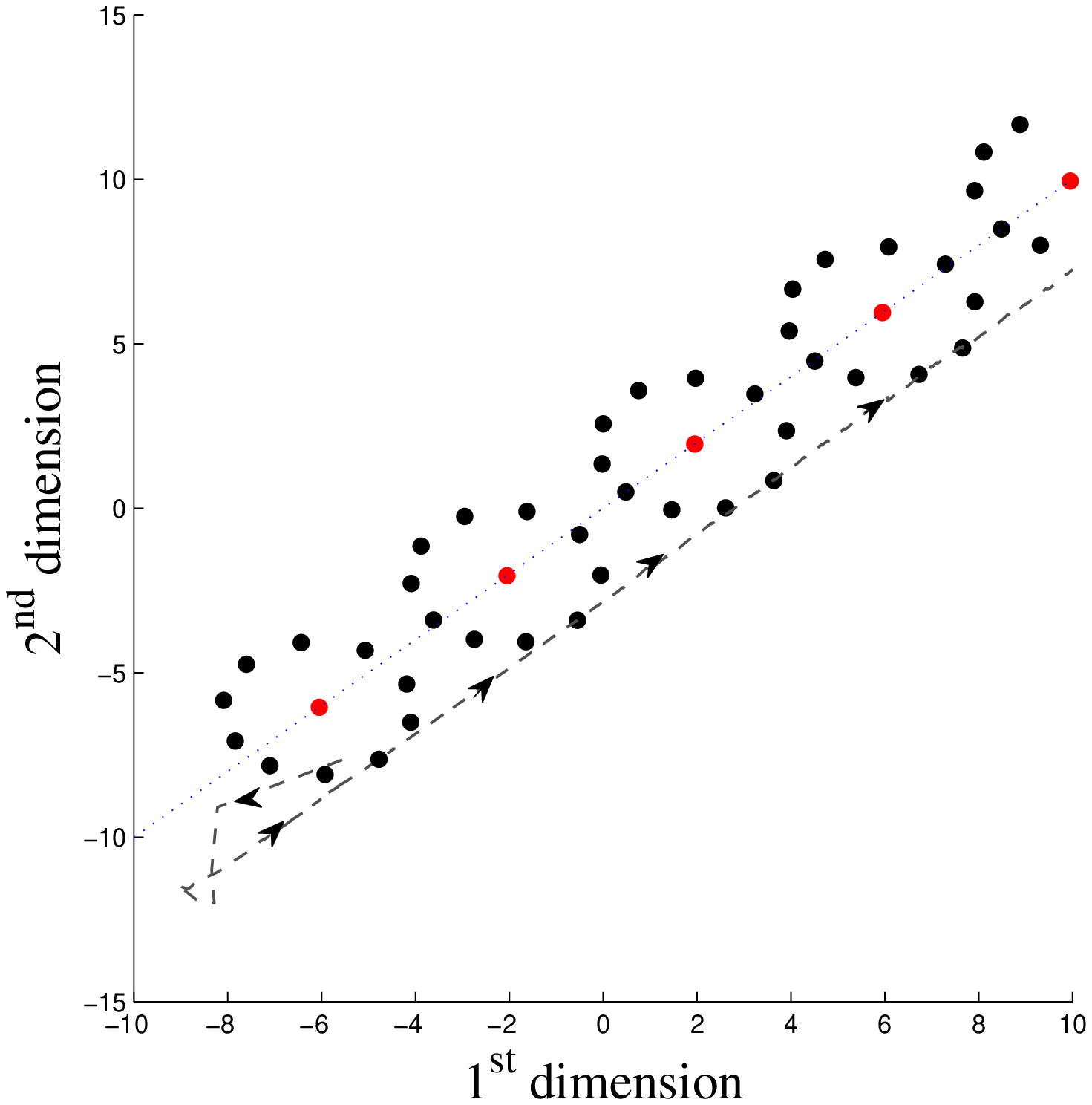}
                \label{fig::trajecotry1}
        }%
        \subfloat[]{
                \includegraphics[width=0.45\textwidth]{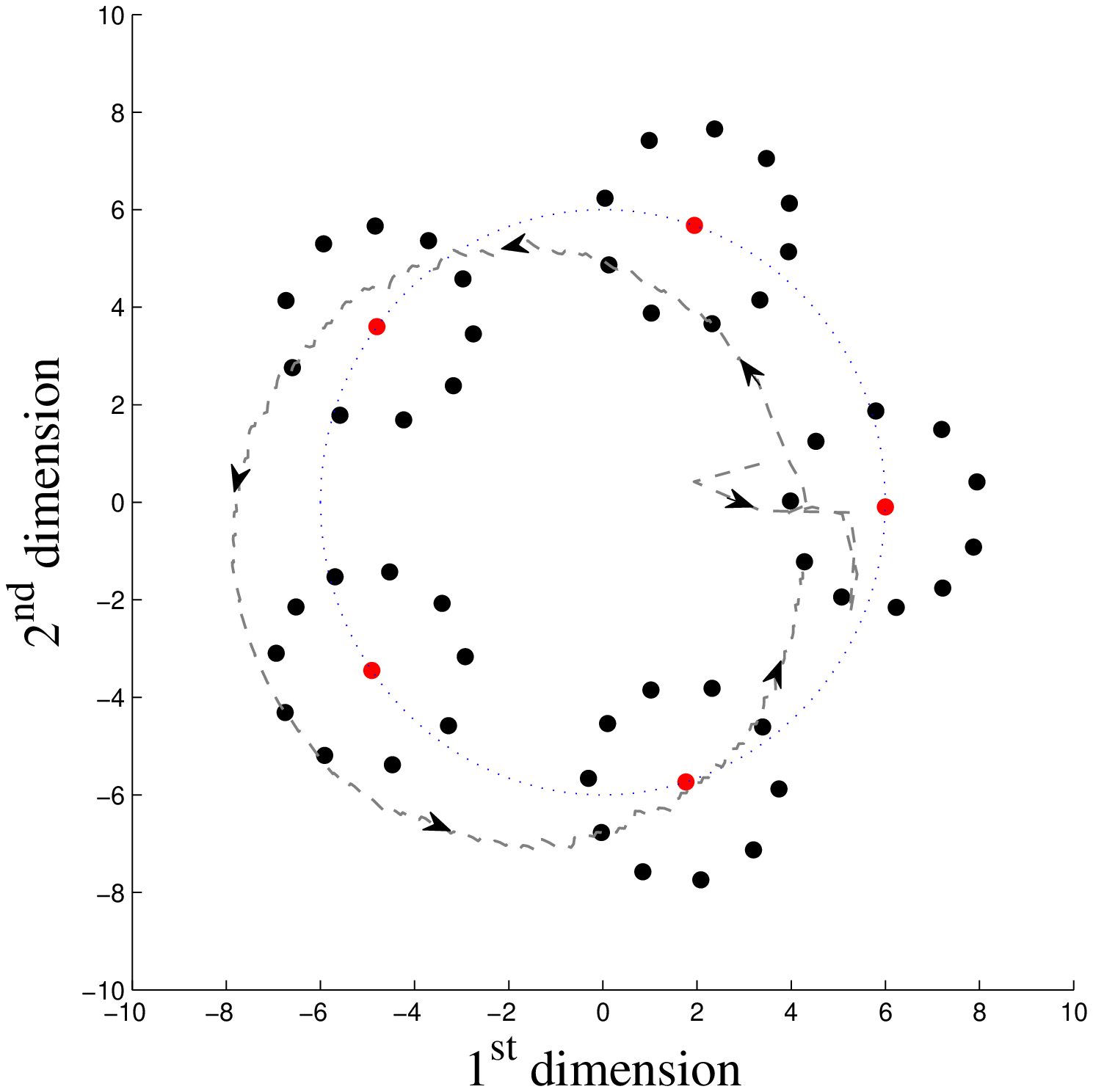}
                \label{fig::trajectory2}
        }
        \caption{Trajectory of an agent during tracking the center which is moving in a straight line (\ref{fig::trajecotry1}) and circle (\ref{fig::trajectory2}). }
\end{figure}

\begin{equation}
\vec{x_T}(t)=[C_1(t)+r_0\cos\phi, C_2(t)+r_0\sin\phi]^T,
\end{equation}
where $\vec{C}(t)=[C_1(t),C_2(t)]^T$ is the coordinate of the moving center. For experimental purposes, we considered two types of movements of the center:\\
\begin{itemize}
 \item \textit{Straight Line}\\
 The center moves along the straight line $y=x$ with the initial position $\vec{C}(0)=[−10, −10]^T$ and with the uniform velocity vector $\vec{v_C}(0)=[0.1, 0.1]^T$ . The circle formation of a 10 agent swarm is considered with $k_1=0.1,k_2=2.0,\sigma=3.5, \beta=1.2$, $r=0.1$ and $r_0=2$ The snapshots of the swarms at iterations, $t = 40, 80, 120, 160$, and $200$ are super-imposed in Figure~\ref{track_st_line}. The filled small circles denote the agents and the unfilled circles denote the position of the moving center at that point of time. It is clearly seen from the figure that the formed circle tracks the moving center or in other words the swarm is surrounding the center. In Figure~\ref{fig::trajecotry1} the movement of a single agent during tracking is shown along with the snapshots of the swarm. The trajectory of the agent closely resembles the trajectory of the moving point itself as expected.

\item \textit{Circle}\\
In this case, we consider that the center is moving along the circle $x^2+y^2= 6^2$ with initial position $\vec{C}(0) = [6, 0]^T$ and with a uniform angular velocity such that it completes a full rotation in $T$ iterations. The parameter values of a $10$ agent swarm are taken as $k_1=0.1,k_2=2.0,\sigma=3.5, \beta=1.2, r=0.1, r_0=2$ and $T=200$. We got similar results as found in the case of the center following a line. The simulation results are shown in Figure~\ref{track_circular}. The Figure~\ref{fig::trajectory2} shows the trajectory of a single agent during tracking for circular motion of the moving point.
\end{itemize}

\section{Conclusion}
This article proposed a simple but elegant method to form elementary geometric structure with multi-agent system. This method is partly inspired from a foraging dynamics where the agents communicate with each other to accomplish a common goal. The attractant-repellent profile and the convergence point of the dynamics is modified using a parametric target vector and an extra repulsion term among the nearest neighbours to meet the demand of target-shape and nearly even density of agents over it. 

Only simple geometric structures like straight line, ellipse, circle and square have been considered to keep the understanding of the swarm-mechanism easier and vivid. The tracking ability has also been demonstrated point moving along a straight line or around a circle and surrounding it with agents in desired shape. The future work in this field can extend to the formation of more complex geometric structures with their Cartesian or polar equation supplied. The emergence of deterministic chaos in the perspective of shape formation can also be investigated.

\bibliographystyle{IEEEtran} 
\bibliography{ref}

\begin{thebibliography}{10}
\providecommand{\url}[1]{#1}
\csname url@samestyle\endcsname
\providecommand{\newblock}{\relax}
\providecommand{\bibinfo}[2]{#2}
\providecommand{\BIBentrySTDinterwordspacing}{\spaceskip=0pt\relax}
\providecommand{\BIBentryALTinterwordstretchfactor}{4}
\providecommand{\BIBentryALTinterwordspacing}{\spaceskip=\fontdimen2\font plus
\BIBentryALTinterwordstretchfactor\fontdimen3\font minus
  \fontdimen4\font\relax}
\providecommand{\BIBforeignlanguage}[2]{{%
\expandafter\ifx\csname l@#1\endcsname\relax
\typeout{** WARNING: IEEEtran.bst: No hyphenation pattern has been}%
\typeout{** loaded for the language `#1'. Using the pattern for}%
\typeout{** the default language instead.}%
\else
\language=\csname l@#1\endcsname
\fi
#2}}
\providecommand{\BIBdecl}{\relax}
\BIBdecl

\bibitem{Buhletal06}
J.~Buhl, D.~J.~T. Sumpter, I.~D. Couzin, J.~J. Hale, E.~Despland, E.~R. Miller,
  and S.~J. Simpson, ``From disorder to order in marching locusts,''
  \emph{Science}, vol. 312, no. 5778, pp. 1402--1406, 2006.

\bibitem{egerstedt01}
M.~Egerstedt and X.~Hu, ``Formation constrained multi-agent control,''
  \emph{Robotics and Automation, IEEE Transactions on}, vol.~17, no.~6, pp.
  947--951, Dec 2001.

\bibitem{barnesetal09}
L.~Barnes, M.-A. Fields, and K.~Valavanis, ``Swarm formation control utilizing
  elliptical surfaces and limiting functions,'' \emph{Systems, Man, and
  Cybernetics, Part B: Cybernetics, IEEE Transactions on}, vol.~39, no.~6, pp.
  1434--1445, Dec 2009.

\bibitem{fm04a}
J.~Fax and R.~Murray, ``Information flow and cooperative control of vehicle
  formations,'' \emph{Automatic Control, IEEE Transactions on}, vol.~49, no.~9,
  pp. 1465--1476, Sept 2004.

\bibitem{buzogany}
P.~M. Buzogany, E. and J.~d’Azzo, ``Automated control of aircraft in
  formation flight,'' in \emph{Proc. AIAA Conf. Guidance, Navigation, and
  Control}, p. 1349–1370.

\bibitem{yamaguchi}
H.~Yamaguchi, T.~Arai, and G.~Beni, ``A distributed control scheme for multiple
  robotic vehicles to make group formations,'' \emph{Robotics and Autonomous
  Systems}, vol.~36, no.~4, pp. 125 -- 147, 2001.

\bibitem{Lin}
J.~Lin, K.~Hwang, and Y.~Wang, ``A simple scheme for formation control based on
  weighted behavior learning,'' \emph{Neural Networks and Learning Systems,
  IEEE Transactions on}, vol.~25, no.~6, pp. 1033--1044, June 2014.

\bibitem{ZelinskiKS03}
S.~Zelinski, T.-J. Koo, and S.~Sastry, ``Optimization-based formation
  reconfiguration planning for autonomous vehicles.'' in \emph{ICRA}.\hskip 1em
  plus 0.5em minus 0.4em\relax IEEE, 2003, pp. 3758--3763.

\bibitem{1243403}
R.~Saber, W.~Dunbar, and R.~Murray, ``Cooperative control of multi-vehicle
  systems using cost graphs and optimization,'' in \emph{American Control
  Conference, 2003. Proceedings of the 2003}, vol.~3, June 2003, pp. 2217--2222
  vol.3.

\bibitem{coordination}
L.~Chaimowicz and V.~Kumar, ``\BIBforeignlanguage{English}{Aerial shepherds:
  Coordination among uavs and swarms of robots},'' in
  \emph{\BIBforeignlanguage{English}{Distributed Autonomous Robotic Systems
  6}}, R.~Alami, R.~Chatila, and H.~Asama, Eds.\hskip 1em plus 0.5em minus
  0.4em\relax Springer Japan, 2007, pp. 243--252.

\bibitem{lawton}
J.~Lawton, R.~Beard, and B.~Young, ``A decentralized approach to formation
  maneuvers,'' \emph{Robotics and Automation, IEEE Transactions on}, vol.~19,
  no.~6, pp. 933--941, Dec 2003.

\bibitem{balch}
T.~Balch and R.~Arkin, ``Behavior-based formation control for multirobot
  teams,'' \emph{Robotics and Automation, IEEE Transactions on}, vol.~14,
  no.~6, pp. 926--939, Dec 1998.

\bibitem{campout}
T.~Huntsberger, P.~Pirjanian, A.~Trebi-Ollennu, H.~Das~Nayar, H.~Aghazarian,
  A.~Ganino, M.~Garrett, S.~Joshi, and P.~Schenker, ``Campout: a control
  architecture for tightly coupled coordination of multirobot systems for
  planetary surface exploration,'' \emph{Systems, Man and Cybernetics, Part A:
  Systems and Humans, IEEE Transactions on}, vol.~33, no.~5, pp. 550--559, Sept
  2003.

\bibitem{leadfollow}
L.~Consolini, F.~Morbidi, D.~Prattichizzo, and M.~Tosques, ``Leader-follower
  formation control of nonholonomic mobile robots with input constraints,''
  \emph{Automatica}, vol.~44, no.~5, pp. 1343--1349, 2008.

\bibitem{1570486}
L.~Chaimowicz, N.~Michael, and V.~Kumar, ``Controlling swarms of robots using
  interpolated implicit functions,'' in \emph{Robotics and Automation, 2005.
  ICRA 2005. Proceedings of the 2005 IEEE International Conference on}, April
  2005, pp. 2487--2492.

\bibitem{1180220}
K.~Fujibayashi, S.~Murata, K.~Sugawara, and M.~Yamamura, ``Self-organizing
  formation algorithm for active elements,'' in \emph{Reliable Distributed
  Systems, 2002. Proceedings. 21st IEEE Symposium on}, 2002, pp. 416--421.

\bibitem{1205192}
A.~Jadbabaie, J.~Lin, and A.~Morse, ``Coordination of groups of mobile
  autonomous agents using nearest neighbor rules,'' \emph{Automatic Control,
  IEEE Transactions on}, vol.~48, no.~6, pp. 988--1001, June 2003.

\bibitem{4209428}
M.~Hsieh, S.~Loizou, and V.~Kumar, ``Stabilization of multiple robots on stable
  orbits via local sensing,'' in \emph{Robotics and Automation, 2007 IEEE
  International Conference on}, April 2007, pp. 2312--2317.

\bibitem{4915742}
L.~Barnes, M.-A. Fields, and K.~Valavanis, ``Swarm formation control utilizing
  elliptical surfaces and limiting functions,'' \emph{Systems, Man, and
  Cybernetics, Part B: Cybernetics, IEEE Transactions on}, vol.~39, no.~6, pp.
  1434--1445, Dec 2009.

\bibitem{mariottini}
G.~Mariottini, F.~Morbidi, D.~Prattichizzo, G.~Pappas, and K.~Daniilidis,
  ``Leader-follower formations: Uncalibrated vision-based localization and
  control,'' in \emph{Robotics and Automation, 2007 IEEE International
  Conference on}, April 2007, pp. 2403--2408.

\bibitem{shao}
J.~Shao, G.~Xie, and L.~Wang, ``Leader-following formation control of multiple
  mobile vehicles,'' \emph{Control Theory Applications, IET}, vol.~1, no.~2,
  pp. 545--552, March 2007.

\bibitem{desai02}
J.~P. Desai, ``A graph theoretic approach for modeling mobile robot team
  formations,'' \emph{J. Field Robotics}, vol.~19, no.~11, pp. 511--525, 2002.

\bibitem{fredslund02}
J.~Fredslund and M.~J. Mataric, ``A general algorithm for robot formations
  using local sensing and minimal communication,'' \emph{IEEE T. Robotics and
  Automation}, vol.~18, no.~5, pp. 837--846, 2002.

\bibitem{pilz}
U.~Pilz and H.~Werner, ``Convergence speed in formation control of multi-agent
  systems - a robust control approach,'' in \emph{Decision and Control (CDC),
  2013 IEEE 52nd Annual Conference on}, Dec 2013, pp. 6067--6072.

\bibitem{gefua}
S.~S. Ge, C.-H. Fua, and K.-W. Lim, ``Multi-robot formations: queues and
  artificial potential trenches,'' in \emph{Robotics and Automation, 2004.
  Proceedings. ICRA '04. 2004 IEEE International Conference on}, vol.~4, April
  2004, pp. 3345--3350 Vol.4.

\bibitem{gazi}
V.~Gazi, ``Swarm aggregations using artificial potentials and sliding-mode
  control,'' \emph{Robotics, IEEE Transactions on}, vol.~21, no.~6, pp.
  1208--1214, Dec 2005.

\bibitem{kim06}
D.~Kim, H.~Wang, G.~Ye, and S.~Shin, ``Decentralized control of autonomous
  swarm systems using artificial potential functions: analytical design
  guidelines,'' in \emph{Decision and Control, 2004. CDC. 43rd IEEE Conference
  on}, vol.~1, Dec 2004, pp. 159--164 Vol.1.

\bibitem{leonard01}
N.~Leonard and E.~Fiorelli, ``Virtual leaders, artificial potentials and
  coordinated control of groups,'' in \emph{Decision and Control, 2001.
  Proceedings of the 40th IEEE Conference on}, vol.~3, 2001, pp. 2968--2973
  vol.3.

\bibitem{zelinski}
S.~Zelinski, T.-J. Koo, and S.~Sastry, ``Optimization-based formation
  reconfiguration planning for autonomous vehicles,'' in \emph{ICRA}, 2003, pp.
  3758--3763.

\bibitem{fua07}
C.-H. Fua, S.~Ge, K.~D. Do, and K.-W. Lim, ``Multirobot formations based on the
  queue-formation scheme with limited communication,'' \emph{Robotics, IEEE
  Transactions on}, vol.~23, no.~6, pp. 1160--1169, Dec 2007.

\bibitem{tucker}
T.~Balch and M.~Hybinette, ``Social potentials for scalable multi-robot
  formations,'' in \emph{Robotics and Automation, 2000. Proceedings. ICRA '00.
  IEEE International Conference on}, vol.~1, 2000, pp. 73--80 vol.1.

\bibitem{kelly}
K.~Kelly, ``Out of control: the new biology of machines, social systems and the
  economic world,'' 1995.

\bibitem{fukuda}
T.~Fukuda, S.~Nakagawa, Y.~Kawauchi, and M.~Buss, ``Structure decision method
  for self organising robots based on cell structures-cebot,'' in
  \emph{Robotics and Automation, 1989. Proceedings., 1989 IEEE International
  Conference on}, May 1989, pp. 695--700 vol.2.

\bibitem{Shen}
W.-M. Shen, P.~Will, A.~Galstyan, and C.-M. Chuong,
  ``\BIBforeignlanguage{English}{Hormone-inspired self-organization and
  distributed control of robotic swarms},''
  \emph{\BIBforeignlanguage{English}{Autonomous Robots}}, vol.~17, no.~1, pp.
  93--105, 2004.

\bibitem{turing}
A.~Turing, ``The chemical basis of morphogenesis,'' 1952, pp. 37--72.

\bibitem{taylor}
T.~Taylor, ``A genetic regulatory network-inspired real-time controller for a
  group of underwater robots.''

\bibitem{guo}
H.~Guo, Y.~Meng, and Y.~Jin, ``Swarm robot pattern formation using a
  morphogenetic multi-cellular based self-organizing algorithm,'' in
  \emph{Robotics and Automation (ICRA), 2011 IEEE International Conference on},
  May 2011, pp. 3205--3210.

\bibitem{bioinsp}
K.~Yeom, ``Bio-inspired automatic shape formation for swarms of
  self-reconfigurable modular robots,'' in \emph{Bio-Inspired Computing:
  Theories and Applications (BIC-TA), 2010 IEEE Fifth International Conference
  on}, Sept 2010, pp. 469--476.

\bibitem{fierro02}
J.~R. Spletzer, A.~K. Das, R.~Fierro, C.~J. Taylor, V.~Kumar, and J.~P.
  Ostrowski, ``Cooperative localization and control for multi-robot
  manipulation,'' in \emph{IROS}, 2001, pp. 631--636.

\bibitem{das2002}
A.~K. Das, R.~Fierro, V.~Kumar, J.~P. Ostrowski, J.~Spletzer, and C.~J. Taylor,
  ``A vision-based formation control framework,'' \emph{Robotics and
  Automation, IEEE Transactions on}, vol.~18, no.~5, pp. 813--825, 2002.

\bibitem{dunbarmurry02}
W.~Dunbar and R.~Murray, ``Model predictive control of coordinated
  multi-vehicle formations,'' in \emph{Decision and Control, 2002, Proceedings
  of the 41st IEEE Conference on}, vol.~4, Dec 2002, pp. 4631--4636 vol.4.

\bibitem{kobayashi}
T.~N. Kobayashi, F. and F.~Kojima, ``Reformation of mobile robots using genetic
  algorithm and reinforcement learning,'' in \emph{In SICE Annu. Conf, pages
  2902–2907}, 2003.

\bibitem{hirota}
K.~Hirota, T.~Kuwabara, K.~Ishida, A.~Miyanohara, H.~Ohdachi, T.~Ohsawa,
  W.~Takeuchi, N.~Yubazaki, and M.~Ohtani, ``Robots moving in formation by
  using neural network and radial basis functions,'' in \emph{Fuzzy Systems,
  1995. International Joint Conference of the Fourth IEEE International
  Conference on Fuzzy Systems and The Second International Fuzzy Engineering
  Symposium., Proceedings of 1995 IEEE Int}, vol.~5, Mar 1995, pp. 91--94
  vol.5.

\bibitem{bonabeauetal99}
E.~Bonabeau, M.~Dorigo, and G.~Theraulaz, \emph{Swarm Intelligence - From
  Natural to Artificial Systems}, ser. Studies in the sciences of
  complexity.\hskip 1em plus 0.5em minus 0.4em\relax Oxford University Press,
  1999.

\bibitem{eberhartetal01}
J.~Kennedy, R.~Eberhart, and Y.~Shi, ``\BIBforeignlanguage{English}{Swarm
  intelligence},'' \emph{\BIBforeignlanguage{English}{The Chemical Educator}},
  vol.~7, no.~2, pp. 123--124, 2002.

\bibitem{gravange}
I.~Gravagne and R.~Marks, ``Emergent behaviors of protector, refugee, and
  aggressor swarms,'' \emph{Systems, Man, and Cybernetics, Part B: Cybernetics,
  IEEE Transactions on}, vol.~37, no.~2, pp. 471--476, April 2007.

\bibitem{andrews}
B.~Andrews, K.~Passino, and T.~Waite, ``Social foraging theory for robust
  multiagent system design,'' \emph{Automation Science and Engineering, IEEE
  Transactions on}, vol.~4, no.~1, pp. 79--86, Jan 2007.

\bibitem{kennedy95}
J.~Kennedy and R.~Eberhart, ``Particle swarm optimization,'' in \emph{Neural
  Networks, 1995. Proceedings., IEEE International Conference on}, vol.~4, Nov
  1995, pp. 1942--1948 vol.4.

\bibitem{delvalleetal08}
Y.~del Valle, G.~K. Venayagamoorthy, S.~Mohagheghi, J.-C. Hernandez, and R.~G.
  Harley, ``Particle swarm optimization: Basic concepts, variants and
  applications in power systems,'' \emph{IEEE Trans. Evolutionary Computation},
  vol.~12, no.~2, pp. 171--195, 2008.

\bibitem{komann08}
M.~Komann, A.~Kr{\"o}ller, C.~Schmidt, D.~Fey, and S.~P. Fekete, ``Emergent
  algorithms for centroid and orientation detection in high-performance
  embedded cameras,'' in \emph{Conf. Computing Frontiers}, 2008, pp. 221--230.

\bibitem{das2014189}
\BIBentryALTinterwordspacing
S.~Das, D.~Goswami, S.~Chatterjee, and S.~Mukherjee, ``Stability and chaos
  analysis of a novel swarm dynamics with applications to multi-agent
  systems,'' \emph{Engineering Applications of Artificial Intelligence},
  vol.~30, no.~0, pp. 189 -- 198, 2014. [Online]. Available:
  \url{http://www.sciencedirect.com/science/article/pii/S0952197613002480}
\BIBentrySTDinterwordspacing

\end{thebibliography}

\end{document}